\documentclass[a4paper, 10pt, conference]{ieeeconf}
\IEEEoverridecommandlockouts
\let\labelindent\relax
\usepackage{enumitem}
\usepackage{cite}
\usepackage{amsmath,amssymb,amsfonts}
\usepackage{algorithmic}
\usepackage{graphicx}
\usepackage{textcomp}
\usepackage{xcolor}
\def\BibTeX{{\rm B\kern-.05em{\sc i\kern-.025em b}\kern-.08em
    T\kern-.1667em\lower.7ex\hbox{E}\kern-.125emX}}

\usepackage[nolist]{acronym}
\usepackage{todonotes}
\usepackage{hyperref}
\usepackage[nameinlink, capitalise]{cleveref}
\usepackage{layouts}
\usepackage{tikz}
\usepackage{pgfplots}
\crefformat{footnote}{#2\footnotemark[#1]#3}

\usepackage{siunitx}      
\usepackage{multirow}     
\usepackage{caption}      
\usepackage{booktabs}     
\usepackage{caption}      
\usepackage{siunitx}      

\begin{document}
\newcommand{\mycomment}[1]{}
\title{\LARGE \bf CARLA-Autoware-Bridge: Facilitating Autonomous Driving Research with a Unified Framework for Simulation and Module Development}

\author{Gemb Kaljavesi$^{1}$, Tobias Kerbl$^{1}$, Tobias Betz$^{1}$, Kirill Mitkovskii$^{2}$ and Frank Diermeyer$^{1}$
\thanks{$^{1}$Gemb Kaljavesi, Tobias Kerbl, Tobias Betz and Frank Diermeyer are with the Institute of Automotive Technology at the Technical University of Munich (TUM), DE-85748 Garching, Germany. {\tt\small\{gemb.kaljavesi, tobias.kerbl, tobias94.betz, diermeyer\}@tum.de}}%
\thanks{$^{2}$Kirill Mitkovski was not supported by any organization. {\tt\small kirill.mitkovskii@gmail.com}}%
}

\mycomment{
\author{\IEEEauthorblockN{1\textsuperscript{st} Gemb Kaljavesi}
\IEEEauthorblockA{\textit{Institute of Automotive Technology } \\
\textit{Technical University of Munich}\\
Garching, Germany \\
gemb.kaljavesi@tum.de}
\and
\IEEEauthorblockN{2\textsuperscript{nd} Tobias Kerbl}
\IEEEauthorblockA{\textit{Institute of Automotive Technology } \\
\textit{Technical University of Munich}\\
Garching, Germany \\
tobias.kerbl@tum.de}
\and
\IEEEauthorblockN{3\textsuperscript{rd} Tobias Betz}
\IEEEauthorblockA{\textit{Institute of Automotive Technology } \\
\textit{Technical University of Munich}\\
Garching, Germany \\
tobi.betz@tum.de}
\and
\IEEEauthorblockN{4\textsuperscript{th} Joel Moriana}
\IEEEauthorblockA{\textit{dept. name of organization (of Aff.)} \\
\textit{name of organization (of Aff.)}\\
City, Country \\
email address or ORCID}
\and
\IEEEauthorblockN{5\textsuperscript{th} Kirill Mitkovskii}
\IEEEauthorblockA{
Stuttgart, Germany \\
kirill.mitkovskii@gmail.com
}
\and
\IEEEauthorblockN{6\textsuperscript{th} Frank Diermeyer}
\IEEEauthorblockA{\textit{Institute of Automotive Technology } \\
\textit{Technical University of Munich}\\
Garching, Germany \\
diermeyer@tum.de}
}
}
    \maketitle

    \begin{abstract}
Extensive testing is necessary to ensure the safety of autonomous driving modules. In addition to component tests, the safety assessment of individual modules also requires a holistic view at system level, which can be carried out efficiently with the help of simulation. Achieving seamless compatibility between a modular software stack and simulation is complex and poses a significant challenge for many researchers. To ensure testing at the system level with state-of-the-art AV software and simulation software, we have developed and analyzed a bridge connecting the CARLA simulator with the AV software Autoware Core/Universe. This publicly available bridge enables researchers to easily test their modules within the overall software. Our investigations show that an efficient and reliable communication system has been established. We provide the simulation bridge as open-source software at
\href{https://github.com/TUMFTM/Carla-Autoware-Bridge}{https://github.com/TUMFTM/Carla-Autoware-Bridge}

\end{abstract}

    \section{Introduction}
\label{sec:introduction}

In 2018, the \ac{who} published a road safety report \cite{worldhealthorganization18} indicating that over one million individuals die annually in road accidents. \acp{av} can potentially decrease this figure \cite{tafidis19}. To realize this potential and facilitate the widespread adoption of autonomous vehicles, they need extensive testing, presenting a significant challenge \cite{koopman16, koopman18}. In general, approaches to autonomous driving can be categorized as modular or end-to-end. Modular approaches involve software comprised of distinct, self-contained modules addressing specific tasks interconnected with each other. In autonomous driving, these modules include mapping, localization, perception, planning and control \cite{yurtsever20}. The complex interdependencies among modules pose a known issue in autonomous robotics literature \cite{yang17}, addressed by frameworks like the \ac{ros} \cite{qui09}. End-to-end approaches aim to treat these modules as a machine learning task, often with a simpler architecture compared to modular ones. 

\begin{figure}[t]
    \centering
    \includegraphics[width=\linewidth]{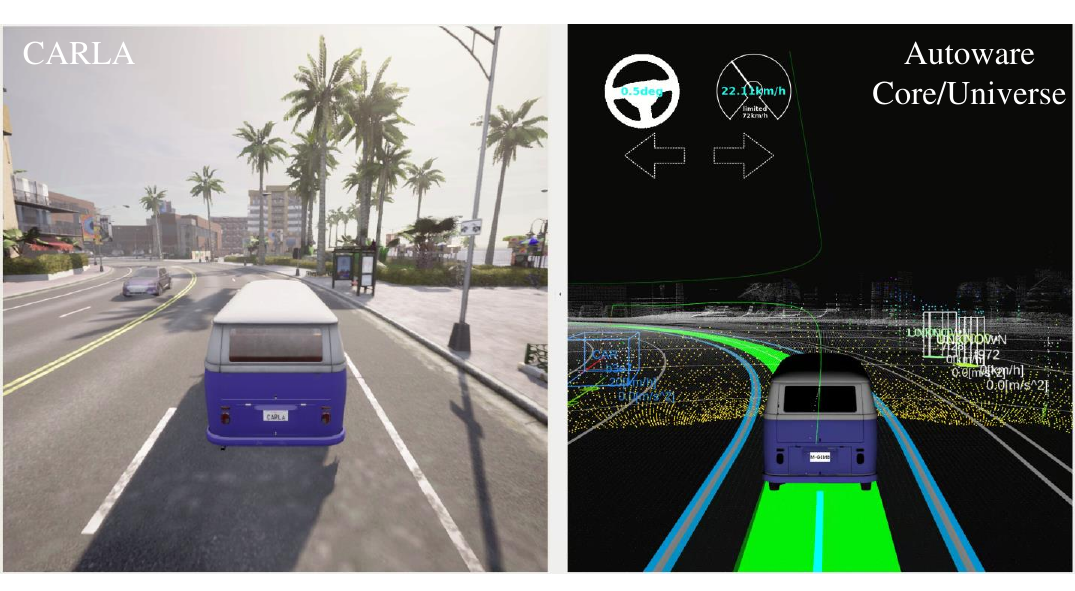}
    \caption{CARLA on the left and Autoware Core/Universe on the right during a simulation.}
    \label{fig:framework}
\end{figure}%

However, modular approaches are more widespread in both industry and research. Their modularity allows teams to focus on specific tasks and independently enhance the system, ensuring operational functionality with practical intermediate outputs \cite{tampuu22}. A modular approach can be tested at both the component and system level. For instance, perception can be evaluated using human-annotated datasets \cite{geiger13}, while planning can be tested through low-fidelity simulations, providing a simplified representation of the environment, the kinematics, and sensors \cite{althoff17}. However, interactions between components are not considered, and not all possible errors in \ac{av} software can be detected \cite{zhong21, advsim21, malay24}. Hence, system-level tests are necessary, involving evaluating the system as a whole. This can be efficiently achieved through \ac{hf} simulations, which incorporate mathematical models of the environment, the \ac{av}, surrounding road users, and sensors. 

However, testing self-developed modules within a comprehensive \ac{hf} simulation framework is currently not possible with state-of-the-art \ac{av} software, nor are detailed performance analyses available for older frameworks.

\subsection{Related Work}
There is an increasing number of publicly available software projects in the field of autonomous driving. Researchers and the automotive industry can profit at a large scale from these open-source projects, sharing their results and leveraging the potential of a drastically growing community \cite{yurtsever20}.

\subsubsection{Open-Source Software for Autonomous Vehicles}
Currently, the most advanced modular open-source software stacks for autonomous driving are Autoware Core/Universe \cite{git_autoware} and Apollo \cite{git_apollo}. Autoware is developed and maintained by the Autoware Foundation, consisting of more than 70 international members from academia and industry \cite{aw_homepage}. Whereas its first version, Autoware.AI \cite{git_auowareai}, is \ac{ros}\,1 based, the latest distribution, Autoware Core/Universe, is based on \ac{ros}\,2. Baidu is the company responsible for the development of Apollo. It was first released in 2017, and its latest update, Apollo 9.0, was launched at the end of 2023. In addition to Baidu, the Apollo project comprises more than 100 members, mainly from industry \cite{apollo_homepage}. It is also to mention that besides \ac{av} software with modular architectures, there are numerous end-to-end approaches for autonomous driving, such as ReasonNet \cite{sha23}. As this work focuses on modular \ac{av} software, these are not considered further. Due to the widespread usage of \ac{ros} within the research community and the strong focus on open-source, Autoware constitutes a good option for research purposes. Apollo, however, lacks proper documentation and focuses more on industrial-scale in-field testing and data collection.

\subsubsection{High-Fidelity Simulators for Autonomous Vehicles}
Testing \acp{av} with \ac{hf} simulators complements real-world testing and enables a more precise prediction of the \ac{av}'s performance in the real world. The \ac{hf} simulation data must closely match the real-world data \cite{din23, zho21}. Important performance indicators for \ac{hf} simulation frameworks are the number of available assets (e.g., maps, sensors, or vehicles), scenario design and execution capabilities, and compatibility with external systems \cite{zho21, ros19}.
There are various commercial \ac{hf} simulators (e.g., CarMaker \cite{carmaker}) available, and a lot of companies involved in \ac{av} development built up their own simulation frameworks, which are not publicly available \cite{zho21}. However, there are also various open-source projects that aim to provide \ac{hf} simulation for research purposes. In order to evaluate the feasibility of different open-source \ac{hf} simulators for our purpose, a subset of criteria partially based on \cite{ros19, zho21, din23} is used. Discontinued projects, like the widespread LGSVL simulator \cite{guo20}, are not suitable since they do not undergo further development. Some simulators are tailored for specific use cases, 
while others lack of realism (e.g., esmini \cite{git_esmini}). Since a \ac{hf} simulator for autonomous driving in real-world settings is required, these simulators are also not an option. Both CARLA \cite{dos17} and AWSIM \cite{git_awsim} fulfill these initial requirements. They include sensor models for LiDAR, Camera, IMU, GNSS, and vehicle status information. Whereas CARLA comes with several different vehicle models and maps, AWSIM only includes one of each. In general, CARLA outperforms AWSIM in terms of provided assets. Both provide a \ac{ros} interface for communication with external \ac{av} software. As AWSIM was developed specifically for Autoware Core/Universe, no further middleware is required to use AWSIM with the latest Autoware release. For the end-of-life Autoware.AI distribution,  options for closed-loop testing with CARLA or the LGSVL simulator were implemented \cite{git_autowareaisim}. A more recent bridge enables the simulation of Apollo 8.0 within CARLA \cite{git_carlaapollo}. Yet, there is no possibility of leveraging the extensive assets provided by CARLA for testing and development of the prominent Autoware Core/Universe distribution. 

\subsection{Contribution}
To provide a valuable option of system-level testing in a \ac{hf} simulation environment with extensive assets, especially for research purposes, this work introduces a bridge to link two widely adopted open-source projects:  CARLA as simulation environment and Autoware Core/Universe as state-of-the-art software under test. Further, experiments were conducted to show the feasibility of the framework, namely the CARLA-Autoware-Bridge, for closed-loop simulation at system-level.

The reminder of this work is structured as follows. \cref{sec:framework} provides a detailed description of the proposed simulation framework. \cref{sec:experiments} explains the design of the conducted experiments to showcase the capabilities of the CARLA-Autoware-Bridge. Results obtained from the experiments are presented in \cref{sec:results}. \cref{sec:conclusion} derives conclusions from the work, which we further discuss to indicate future research directions.

    \section{Unified Simulation Framework}
\label{sec:framework}
We outline our framework requirements in Sec. \ref{ssec:req}. Following that, we provide an overview of the framework in interaction with simulation and \ac{av} software in Sec. \ref{ssec:over}. Subsequently, we detail the data exchange from CARLA to Autoware in Sec. \ref{ssec:ctoaw}, as well as Autoware to CARLA in Sec. \ref{ssec:awtoc}. Finally, Sec. \ref{ssec:aw_deps} presents the Autoware dependencies that have been developed.

\subsection{Requirements}
\label{ssec:req}
To accomplish a \ac{hf} simulation in real-time and offer an easy-to-use framework, we defined requirements to be fulfilled by the CARLA-Autoware-Bridge. As its main functionality, the bridge must enable the bidirectional communication between CARLA and Autoware to transfer sensor data and control commands \textbf{(RQ-1)}. The sensor setup and the vehicle model simulated in CARLA must correspond to the setup Autoware expects \textbf{(RQ-2)}. To maintain CARLA's real-time simulation capabilities, the bridge must operate efficiently. It should minimize computational effort and keep introduced latency to a minimum \textbf{(RQ-3)}. The framework should quickly adapt to diverse sensor setups and vehicle models, ensuring a fast and easy-to-use configuration process \textbf{(RQ-4)}.

\begin{itemize}[leftmargin=1.5cm, rightmargin=1cm, itemsep=2pt]
    \item[\textbf{RQ-1}] \textbf{Interoperability:} The bridge should offer an interface between the latest releases of CARLA and Autoware Core/Universe.
    \item[\textbf{RQ-2}] \textbf{Compatibility:} CARLA and Autoware should both be provided with a consistent sensor and vehicle setup.
    \item[\textbf{RQ-3}] \textbf{Efficiency:} The bridge should have a low overall latency and minimal hardware requirements.
    \item[\textbf{RQ-4}] \textbf{Adaptability:} Sensor setup and vehicle model should be configurable in a fast and flexible manner.
\end{itemize}

\begin{figure*}[h]
\includegraphics[width=\textwidth]{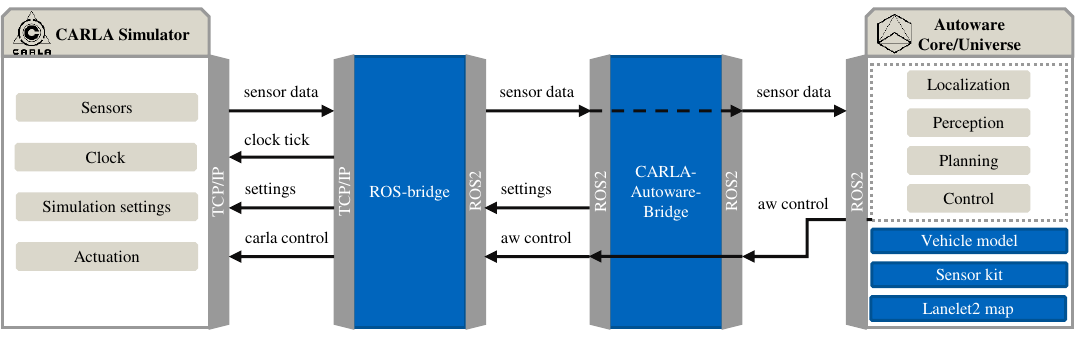}\caption{Overview of the framework for system-level testing with the components developed or customized as part of in this work highlighted in blue.}
\label{fig:release}
\end{figure*}%

\subsection{Overview}
\label{ssec:over}
\cref{fig:release} depicts the software framework for system-level simulation of Autoware Core/Universe using CARLA. The CARLA simulator, with some of its functions, is shown on the left. It can execute control commands through a simulated actuator, save and implement simulation settings (e.g., weather), manage and publish the simulation time, and generate diverse sensor data. Communication to and from the CARLA simulator occurs via TCP/IP.

The framework leverages the existing \ac{ros}-Bridge for the CARLA Simulator \cite{git_rosbridge}, which has been enhanced to meet the authors' requirements. The primary role of the \ac{ros}-Bridge is to facilitate communication between TCP/IP and \ac{ros}\,2. It also provides the server with the next time step after receiving all expected sensor data, ensuring synchronization and reproducibility of the simulation.

The CARLA-Autoware-Bridge, developed as part of this work, links the \ac{av} software and the \ac{ros}-Bridge. It manages initial simulation settings, sensor configurations of the \ac{av}, and the \ac{av} itself. Additionally, it converts sensor information from CARLA if necessary. For efficiency, non-converted sensor data is remapped to the correct topics and sent directly from the \ac{ros}-Bridge to Autoware. Control data sent by Autoware is transmitted directly to the \ac{ros}-Bridge and converted into CARLA control commands using a controller developed as part of this work. 

Furthermore, we created a vehicle model, an exemplary sensor kit, and a Lanelet2 map required for Autoware in this context. This framework makes autonomous driving possible with a specific CARLA vehicle model, on one particular map, and with any sensor setup within the simulation.

\mycomment{
\begin{table}[h]
    \centering
    \label{tab:interface}
     \caption{Description of the \ac{ros} Interfaces of CARLA and Autoware}
    \begin{tabularx}{\linewidth}{p{0.45\linewidth} p{0.45\linewidth}}
        \toprule
        \textbf{CARLA} & \textbf{Autoware} \\
        \midrule
        \textbf{Sensor Data} & \\
        \midrule
        CarlaEgoVehicleStatus & \\
        Odometry & \\
        Camera & \\
        Lidar & \\
        Gnss & \\
        IMU & \\
        & SteeringReport \\
        & VelocityReport \\
        & PoseWithCovarianceStamped \\
        \textbf{Control Commands} & \\
        \midrule
        AckermannDrive & AckermannControlCommand \\
        \bottomrule
    \end{tabularx}
\end{table}
}
\subsection{CARLA to Autoware: Sensor Data}
\label{ssec:ctoaw}
CARLA supports a variety of sensors, but its default data output is incompatible with Autoware Core/Universe. To optimize efficiency, the \ac{ros}-Bridge has been modified to handle and modify extensive data, such as camera images or LiDAR point clouds. As a result, that data does not need to pass through the CARLA-Autoware-Bridge. Adjustments made in either the \ac{ros}-Bridge or the CARLA-Autoware-Bridge include modifying the quality of service for messages and converting data projection from a left-handed coordinate system to a right-handed coordinate system. Additionally, certain information from CARLA needed summarization, while others required splitting. Moreover, \ac{ros}\,2 messages had to be converted into messages suitable for Autoware Core/Universe.

\subsection{Autoware to CARLA: Control Commands}
\label{ssec:awtoc}
The Autoware Core/Universe Controller transmits an Ackermann Control Command, encompassing target speed, target acceleration, and required tire angle, among other parameters. Autoware manages the vehicle's longitudinal movement through a \ac{pid} controller and the lateral movement using a \ac{mpc}.
However, direct transfer of the Autoware Control Command to the CARLA simulation is not feasible. CARLA accepts the gas pedal position, brake pedal position, and steering wheel angle as inputs. To bridge this gap, we fine-tuned a \ac{pid} controller to convert the longitudinal commands. The \ac{ros}-Bridge incorporates a simple \ac{pid} controller implementation for this purpose.
In addition to that, we generated a mapping to convert the tire angle to a steering angle. To enhance framework usability and ensure compatibility with future Autoware Core/Universe versions, controller parameter tuning for the Autoware \ac{pid} or \ac{mpc} was omitted.

\subsection{Autoware Dependencies}
\label{ssec:aw_deps}
In addition to ensuring compatible communication between Autoware and CARLA, providing the necessary dependencies for Autoware is crucial. Firstly, Autoware requires the simulated vehicle as a 3D model for rendering, along with the associated vehicle parameters. To achieve this, the \textit{VW T2 2021} was extracted from CARLA, converted into the correct format, and measured using the open-source tool Blender \cite{git_blender}.

Additionally, Autoware requires the sensor kit, which provides information about the type and position of sensors within the simulation. For this purpose, an exemplary sensor kit was developed for the vehicle, serving as a template that can be expanded to include additional sensors.

Finally, Autoware needs an HD map of the CARLA map in Lanelet2\cite{lanelet2} format. The available CARLA maps, originally in OpenDrive format \cite{dupuis2010opendrive}, were converted with the assistance of the CommonRoad Scenario Designer\cite{scenariodesigner}. Furthermore, any missing or incorrect information in the maps was subsequently edited by hand.

    \section{Experiment Design}
\label{sec:experiments}
We conducted experiments to demonstrate the capabilities of the CARLA-Autoware-Bridge and validate whether we satisfy the requirements specified in \cref{ssec:veh_exp}. For all experiments, we used the latest version of Autoware Core/Universe at the time of publication and CARLA 0.9.15. In the subsequent sections, we outline the design of these experiments. First, \cref{ssec:veh_exp} details experiments demonstrating the accurate transmission of control commands. Secondly, we defined test cases in  \cref{ssec:perf_exp} to benchmark the overall performance of the simulation framework in terms of computational efficiency. 
All the experiments are performed on a simulation workstation equipped with an Intel i9-14900K CPU (up to \SI{6.0}{\giga\hertz}), \SI{192}{\giga\byte} RAM, and an NVIDIA RTX 4090 GPU. 

\subsection{Transmission of Control Commands}
\label{ssec:veh_exp}
We designed these experiments to evaluate our pipeline to transfer control commands from Autoware to CARLA, which is outlined in \cref{ssec:ctoaw}. To enable a separate evaluation of both the introduced \ac{pid} controller and the steering angle mapping, the experiment is divided into a longitudinal and lateral test case. This division allows for an isolated observation of the performance of each introduced module. At the beginning of each test run, the vehicle is placed in the same initial pose on the map \textit{Town10HD} and is assigned either the goal position for the longitudinal or lateral case.

The \ac{pid} controller for longitudinal control is evaluated on a straight, with Autoware given the goal to drive along it with a constant target velocity of \SI{8.33}{\meter\per\second}. We track the actual velocity of the vehicle and compare it against the given target velocity. The velocity curve is evaluated qualitatively regarding potential over- or undershooting and target value tracking. The experiment is conducted once with the default \ac{pid} settings of the \ac{ros}-Bridge and once with manually tuned parameters obtained within this work. 

The mapping of the steering angle for lateral control is assessed in a scenario in which the vehicle has to take three consecutive sharp turns. The experiment is performed once before the introduction of the mapping and afterward. To compare these two variants,  we track the lateral deviation from the center line of the current lane.

\subsection{Performance Benchmark}
\label{ssec:perf_exp}

In this section, we detail the experimental setup designed to evaluate the developed CARLA-Autoware-Bridge. The core objective of our study is to assess the efficiency across various configurations, focusing on the integration of multiple cameras and LiDAR sensors with different resolutions. This setup aims to simulate realistic autonomous driving scenarios, providing insights into the system's processing capabilities, communication efficiency, and overall performance under varying conditions. The benchmarks is designed to cover different performance metrics, including CPU utilization and the latency of the bridge.

For analyzing the performance, we vary the amount of used LiDARs and cameras in CARLA. Ranging from \num{1} to \num{4} LiDARs (\num{1} to \num{3} respectively for cameras), we additionally need to analyse the frames per seconds resulting in the CARLA simulation. Usually creating multiple instances of sensors increase the workload for the simulation environment. For the LiDAR sensor we investigate multiple point densities (\num{1000} pt/s, \num{100000} pt/s, \num{500000}, and \num{1000000} pt/s). For the camera setup we investigate two common resolution for images: \num{1280}p x \num{720}p (HD) and \num{1920}p x \num{1080}p (Full-HD).    

The CPU usage is measured by acquiring the data for the individual processes launched on the compute system with standard Linux \textit{ps}. We sample the process data every \SI{200}{\milli\second}. The processes are the filtered to the relevant once for the \ac{ros}-Bridge and the CARLA-Autoware-Bridge. 

To investigate the latency of the bridge, we use the framework developed by Betz et al. \cite{DAG_Betz.2023}, which is based on \textit{ros2\_tracing} \cite{Bedard.2022}. We measure the latency in the \ac{ros}-Bridge for the individual sensor setups with the corresponding \ac{ros}\,2 callbacks. Note that most of the sensor data are not feed through the CARLA-Autoware-Bridge. Therefore, Autoware is directly subscribing to the modified \ac{ros}-Bridge. 

All the experiments are conducted with the Autoware software running closed-loop. We developed a repeatable scenario according to \cite{betz2023latency} with a fixed start and end goal pose. The scenario is conducted in the map \textit{Town10HD}. In addition to varying and analyzing the LiDAR and camera sensors, the other sensors in the sensor kit were consistently active throughout the test runs.

    \section{Results}
\label{sec:results}
The results of the experiments are presented below. Sec. \ref{ssec:trans_res} includes the findings of the controller transmission analyses, while Sec. \ref{ssec:pb_res} contains the results of the performance benchmarks.

\begin{figure}[t]
  \centering
  \resizebox{\columnwidth}{!}{\input{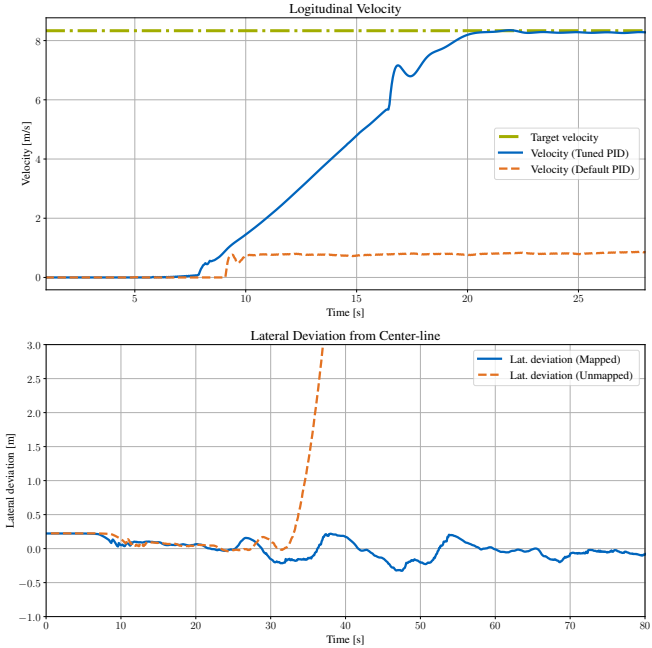}}
  \caption{Comparison of velocity and lateral deviation curves obtained from test runs with and without the introduced \ac{pid} controller and steering angle conversion for control command transmission.}
  \label{fig:devi}
\end{figure}

\subsection{Transmission of Control Commands}
\label{ssec:trans_res}
\Cref{fig:devi} depicts the outcomes of the test runs conducted to assess the conversion and transmission of control commands from Autoware to CARLA. The data from the longitudinal test case reveal that the \ac{pid} Controller from the \ac{ros}-Bridge, equipped with the default parameter set, cannot reach the target velocity. However, with the tuned parameter set, the vehicle achieves its intended target speed, even though the longitudinal velocity shows undesirable deviations in some sections.

The results from the test cases for the lateral control of the vehicle demonstrate that the lateral deviation from the lane center line is generally acceptable with the introduced mapping of the steering commands. The vehicle successfully navigated through all three sharp turns and stayed within the lane at all times. In contrast, the vehicle equipped with the direct forwarding of the steering angles departed from the lane at the first turn, whereby Autoware disengaged.

\subsection{Performance Benchmark}
\label{ssec:pb_res}

As described in section \ref{ssec:perf_exp} we analyse multiple performance metrics to give an insight into the computational efficiency of the developed bridge. 
In \cref{tab:util} the results are depicted for the experiments regarding CPU utilization and the corresponding resulting FPS in the CARLA simulator by varying the sensor setup. 

An increase in the point cloud density directly correlates with a decrease in FPS within CARLA. This outcome is anticipated, as higher point cloud densities demand more computational power to process the number of sensors. Interestingly, a decrease in FPS is accompanied by lower CPU utilization. This suggests that as the simulation becomes less efficient in terms of frame rate, it concurrently requires less computational effort. Furthermore, the addition of multiple sensors to the simulation impacts CPU usage less significant than increasing the point cloud density. This distinction implies that the computational load of processing high-density point clouds is more demanding than managing data from multiple sensors. The FPS drop of CARLA is higher for increased point cloud density as for multiple instances of sensors. For configurations involving a LiDAR setup with up to four sensors and a point cloud density of \num{100000} pt/s, the simulation maintains a FPS rate above \num{30}, indicating a threshold for maintaining real-time performance with acceptable limits. For the camera configurations it is more limited. While ensuring for both resolution a FPS above \num{30}, we cannot add a second camera with Full-HD resolution without dropping the FPS below \num{30}. Rendering camera images is a more resource extensive task compared to LiDAR point cloud generation. However, anomalies in CPU utilization were observed, with occasional slight increase compared to preceding configurations. For example \num{3} LiDARs and \num{1000} pt/s a higher utilization is measured compared to the \num{2} LiDAR setup. These outliers suggest that there may be additional factors at play, influencing CPU demand in less predictable ways. 

The release sensor kit in the open-source repository utilized a single CPU core, demonstrating the ability to achieve an FPS of \num{35} on the used simulation platform. 

\begin{figure}[t!]
  \centering
  \resizebox{\columnwidth}{!}{\input{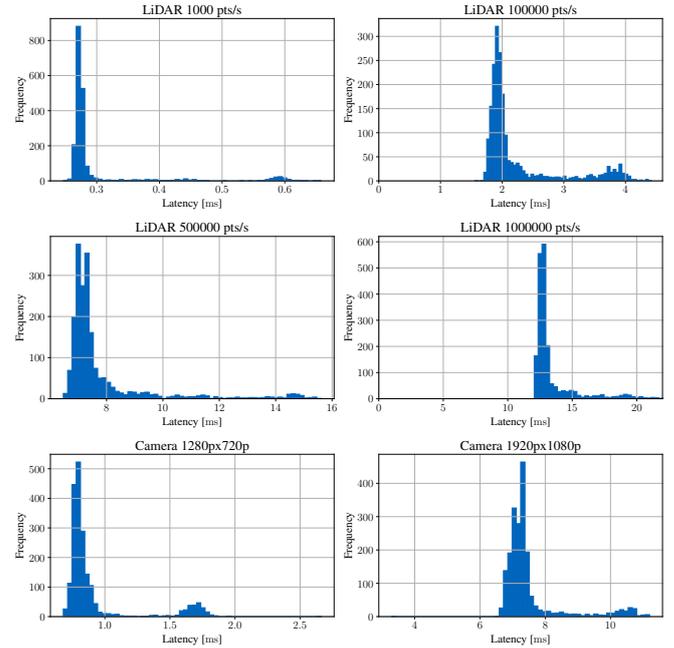}}
  \caption{Histograms displaying the latency of the developed framework, measuring the time between receiving messages from the simulation and outputting messages to the \ac{av} software for various sensors and sensor configurations.}
  \label{fig:histo}
\end{figure}

\begin{table*}[t!]
\centering
\caption{The average CPU utilization and standard deviation, along with the frames per second of the simulation, are presented for various LiDAR-only and camera-only configurations, as well as the released sensor kit. (MCPU=Mean CPU Utilization, DCPU=CPU Utilization Standard Deviation, AFPS=Average Frames per Second).}

\resizebox{\textwidth}{!}{%
\begin{tabular}{|c|c|lll|lll|lll|lll|}
\hline
\multirow{2}{*}{Sensors} & \multirow{2}{*}{Sensor Count} & \multicolumn{3}{c|}{1000 [\SI{}{\text{pts/s}}]} & \multicolumn{3}{c|}{\num{100000} [\SI{}{\text{pts/s}}]} & \multicolumn{3}{c|}{\num{500000} [\SI{}{\text{pts/s}}]} & \multicolumn{3}{c|}{\num{1000000} [\SI{}{\text{pts/s}}]} \\ \cline{3-14}
& & MCPU & DCPU & AFPS & MCPU & DCPU & AFPS & MCPU & DCPU & AFPS & MCPU & DCPU & AFPS \\ \hline

\multirow{4}{*}{LiDAR} & 1 & 135.6 & 0.10 & 101.2 & 132.3 & 2.06 & 95.7 & 109.0 & 1.00 & 52.6 & 98.9 & 1.65 & 36.3\\ \cline{2-14}
& 2 & 125.57 & 3.18 & 96.9 & 120.9 & 3.06 & 78.9 & 92.7 & 2.08 & 35.2 & 90.5 & 1.08 & 21.4\\ \cline{2-14}
& 3 & 127.9 & 1.18 & 87.5 & 117.3 & 2.00 & 61.1 & 93.7 & 1.85 & 25 & 84.4 & 1.22 & 15.1 \\ \cline{2-14}
& 4 & 117.0 & 3.01 & 69.7 & 119.4 & 1.31 & 50.6 & 84.5 & 1.74 & 19.1 & 83.5 & 1.09 & 11.2\\ \cline{1-14}

\multirow{2}{*}{Sensors} & \multirow{2}{*}{Sensor Count} & \multicolumn{3}{c|}{1280p x 720p} & \multicolumn{3}{c|}{1920p x 1080p} \\ \cline{3-8}
& & MCPU & DCPU & AFPS & MCPU & DCPU & AFPS \\ \cline{1-8}

\multirow{3}{*}{Camera} & 1 & 77.1 & 2.04 & 41.0 & 89.1 & 1.50 & 33.2 \\ \cline{2-8}
& 2 & 79.3 & 1.11 & 37.1 & 91.5 & 1.78 & 23.9 \\ \cline{2-8}
& 3 & 65.2 & 0.23 & 23.8 & 91.1 & 3.30 & 18.2 \\ \cline{1-8}

\multirow{2}{*}{Sensors} & \multirow{2}{*}{Sensor Count} & \multicolumn{3}{c|}{Release}\\ \cline{3-5}
& & MCPU & DCPU & AFPS  \\ \cline{1-5}
LiDAR + camera & 2 + 1 & 97.5 & 1.18 & 35.2 & \multicolumn{9}{r}{Release sensor kit: \num{500000}[\SI{}{\text{pts/s}}]; \num{100000}[\SI{}{\text{pts/s}}]; 1280px720p}\\ \cline{1-5}

\end{tabular}%
}

\label{tab:util}
\end{table*}

For the latency metric the results are shown in \cref{fig:histo}. It is observable that the latency increases with the number of points per second for the LiDAR and the camera resolution. The latency increases significantly, with larger point clouds. Although not depicted in the figure, the influence of multiple sensors was also analyzed. It was found that there is no significant impact on latency. The sensor kit released concurrently with this work achieves a maximum latency of less than \SI{15}{\ms} and an average of \SI{7.8}{\ms}.

    \section{Conclusion}
\label{sec:conclusion}
There is a strong demand for a method to test a state-of-the-art modular \ac{av} software stack through \ac{hf} simulation at the system level. With our efforts, we have developed a framework capable of facilitating such system-level testing using the CARLA Simulator and Autoware Core/Universe. We established the communication chain through an enhanced \ac{ros}-Bridge and our CARLA-Autoware-Bridge, also creating a compatible sensor kit, a vehicle model, and a Lanelet2 map. We provided and analyzed a conversion of Ackermann Control Commands to throttle, brake, and steering angle to control the actuators. A performance benchmark was employed to measure the latency and CPU utilization of the bridge, demonstrating its scalability. However, we observed that the bridge's latency increases rapidly with larger LiDAR point clouds, likely due to coordinate changes. This calculation can be implemented more efficiently in the future. Additionally, the map was manually generated; however, future efforts should include an automatic conversion from OpenDrive to Lanelet2 that is compatible with Autoware. In summary, an essential framework for system-level testing of autonomous driving functions has been developed. To facilitate enhanced testing, the system should be expanded in the future with a database of three-dimensional scenarios.
     \section*{Acknowledgment}
Gemb Kaljavesi and Tobias Kerbl, were the main developers of the presented work and main contributer of this paper. Tobias Betz contributed to the performance benchmarks. Kirill Mitkovskii helped with the development of the software. Frank Diermeyer made essential contributions to the conception of the research project and revised the paper critically for important intellectual content. This work was supported by the project Wies’n Shuttle (FKZ 03ZU1105AA) in the MCube cluster, the Federal Ministry of Economic Affairs and Climate Actions within the project ATLAS-L4 (FKZ 19A21048l) and through basic research funds from the Institute for Automotive Technology.
    \begin{acronym}

    \acro{av}[AV]{Automated Vehicle}
    \acro{who}[WHO]{World Health Organization}
    \acro{mpc}[MPC]{Model Predictive Control}
    \acro{pid}[PID]{Proportional-Integral-Derivative}
    \acro{hd}[HD]{High Definition}
    \acroplural{av}[AVs]{Automated Vehicles}
    \acro{ros}[ROS]{Robot Operating System}
    \acro{hf}[HF]{High-Fidelity}
    \acro{lidar}[LiDAR]{Light Detecting and Ranging}

\end{acronym}

    \bibliographystyle{IEEEtran}
    \bibliography{IEEEabrv,bib}

\end{document}